\documentclass[11pt,a4paper]{article}
\usepackage[T1]{fontenc}

\usepackage{flushend}

\usepackage[hyperref]{emnlp2017}

\usepackage{times}
\usepackage{latexsym}

\usepackage{color}
\usepackage{tabularx}
\usepackage{ragged2e}
\newcolumntype{Y}{>{\RaggedRight\let\newline\\\arraybackslash\hspace{0pt}}X} 
\usepackage{multirow}
\usepackage{soul}
\usepackage{amsmath}

\usepackage{graphicx}
\usepackage{pdflscape}
\usepackage{afterpage}
\usepackage{xspace}
\usepackage{ifthen}
\usepackage{amssymb}
\usepackage{amsthm}
\usepackage{arydshln}

\usepackage{standalone}
\usepackage{tkz-berge}
\usepackage{tikz}
\usepackage{xcolor}
\usepackage{pgfplots}
\usepackage[normalem]{ulem}

\usepackage{diagbox}

\newcounter{Lcount}
\newcommand{\squishenum}{
\begin{list}{\arabic{Lcount}. }
{ \usecounter{Lcount}
\setlength{\itemsep}{0pt}
\setlength{\parsep}{0pt}
\setlength{\topsep}{0pt}
\setlength{\partopsep}{0pt}
\setlength{\leftmargin}{2em}
\setlength{\labelwidth}{1.5em}
\setlength{\labelsep}{0.5em} } }

\newcommand{\squishletter}{
\begin{list}{\alph{Lcount}. }
{ \usecounter{Lcount}
\setlength{\itemsep}{0pt}
\setlength{\parsep}{0pt}
\setlength{\topsep}{0pt}
\setlength{\partopsep}{0pt}
\setlength{\leftmargin}{2em}
\setlength{\labelwidth}{1.5em}
\setlength{\labelsep}{0.5em} } }

\newcommand{\squishlist}{
\begin{list}{$\bullet$}
{ \usecounter{Lcount}
\setlength{\itemsep}{0pt}
\setlength{\parsep}{0pt}
\setlength{\topsep}{0pt}
\setlength{\partopsep}{0pt}
\setlength{\leftmargin}{2em}
\setlength{\labelwidth}{1.5em}
\setlength{\labelsep}{0.5em} } }

\newcommand{\squishend}{
\end{list} }

\newtheorem{theorem}{Theorem}[section]

\emnlpfinalcopy 
\def\emnlppaperid{625} 

\definecolor{edited}{HTML}{000000}

\makeatletter
\def\squigglyred{\bgroup \markoverwith{\textcolor{red}{\lower3.5\p@\hbox{\sixly \char58}}}\ULon}
\def\squigglycyan{\bgroup \markoverwith{\textcolor{cyan}{\lower3.5\p@\hbox{\sixly \char58}}}\ULon}
\def\squigglybrown{\bgroup \markoverwith{\textcolor{brown}{\lower3.5\p@\hbox{\sixly \char58}}}\ULon}
\def\squigglypurple{\bgroup \markoverwith{\textcolor{purple}{\lower3.5\p@\hbox{\sixly \char58}}}\ULon}
\def\squigglyblue{\bgroup \markoverwith{\textcolor{blue}{\lower3.5\p@\hbox{\sixly \char58}}}\ULon}
\makeatother

\title{Labeling Gaps Between Words: \\Recognizing Overlapping Mentions with Mention Separators}

\author{Aldrian Obaja Muis \and Wei Lu \\
  Singapore University of Technology and Design \\
  {\tt \{aldrian\_muis,luwei\}@sutd.edu.sg} \\}

\date{}
\begin{document}
\maketitle
\begin{abstract}
In this paper, we propose a new model that is capable of recognizing overlapping mentions.
We introduce a novel notion of {\em mention separators} that can be effectively used to capture how mentions overlap with one another.
On top of a novel {multigraph} representation that we introduce, we show that efficient and exact inference can still be performed.
We present some theoretical analysis on the differences between our model and a recently proposed model for recognizing overlapping mentions, and discuss the possible implications of the differences. Through extensive empirical analysis on standard datasets, we demonstrate the effectiveness of our approach.
\end{abstract}

\section{Introduction}

\begin{figure}
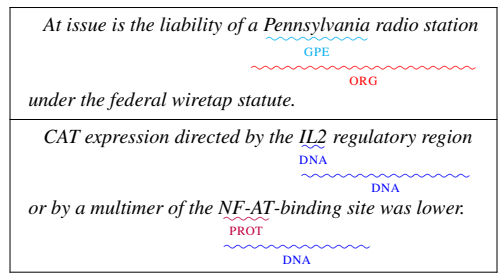

\begin{center}
\scalebox{0.9}
{\footnotesize
\def\arraystretch{0.4}\tabcolsep=8pt
\setlength\doublerulesep{1pt}
\begin{tabular}{|l|}
\hline
\\
{\em \ \ \ At issue is the liability of $\underset{\text{\raisebox{2.5pt}[0pt][0pt]{\color{red}\scriptsize{}\textsc{org}}}}{\squigglyred{\text{a }\vphantom{\_}\underset{\text{\raisebox{2.5pt}[0pt][0pt]{\color{cyan}\scriptsize{}\textsc{gpe}}}}{\squigglycyan{\text{Pennsylvania}}}\text{ radio station}}}$}\\
{\em under the federal wiretap statute.}\\\\
\hline
\\
{\em \ \ \ CAT expression directed by the $\underset{\text{\raisebox{2.5pt}[0pt][0pt]{\scriptsize\color{blue}\textsc{dna}}}}{\squigglyblue{\vphantom{\_}\underset{\text{\raisebox{2.5pt}[0pt][0pt]{\scriptsize\color{blue}\textsc{dna}}}}{\squigglyblue{\text{IL2}}}\text{ regulatory region}}}$}\\
{\em or by a multimer of the $\underset{\text{\raisebox{2.5pt}[0pt][0pt]{\scriptsize\color{blue}\textsc{dna}}}}{\squigglyblue{\vphantom{\_}\underset{\text{\raisebox{2.5pt}[0pt][0pt]{\scriptsize\color{purple}\textsc{prot}}}}{\squigglypurple{\text{NF-AT}}}\text{-binding site}}}$ was lower.}\\
\hline
\end{tabular}
}
\end{center}
\vspace{-2mm}
\caption{Examples of overlapping mentions.}
\vspace{-4mm}
\label{fig:main-example}
\end{figure}

Named entity recognition (NER), or in general the task of recognizing entity mentions\footnote{As noted in \cite{florian2004statistical}, mention recognition is more general than NER, where a mention can be either named, nominal, or pronominal.} in a text, has been a research topic for many years \cite{McCallum:2003,Nadeau:2007,Ratinov:2009,ling2012fine}. 
However, {\color{black} as noted by} \citet{Finkel:2009}, many previous works ignored overlapping mentions, although they are quite common.
Figure \ref{fig:main-example} illustrates some examples of overlapping mentions adapted from existing datasets.
For example, the location mention {\it Pennsylvania} appears within the mention of type organization  {\it a Pennsylvania radio station}.
In practice, overlapping mentions have been found in many existing datasets across different domains \cite{Doddington:2004,Kim:2003,Suominen:2013}.
Developing algorithms that can effectively and efficiently extract overlapping mentions can be crucial for the performance of many downstream tasks such as relation extraction \cite{mintz2009distant,guptatable}, event extraction \cite{lu-roth:2012:ACL2012,li2013joint,nguyen2016joint}, coreference resolution \cite{chang2013constrained,Lu+etal:16a}, question answering \cite{molla2007named}, and equation parsing \cite{Roy:2016}.

{Overlapping mention recognition} is non-trivial, as existing methods that model mention recognition as a sequence prediction problem -- e.g., using linear-chain conditional random fields (CRF) \cite{Lafferty:2001} -- have difficulties in handling overlapping mentions \cite{Alex:2007}.
\citet{Finkel:2009} proposed to use a tree-based constituency parsing model to handle nested entities.\footnote{We note that nested entities are only one of the two kinds of overlapping entities, the other kind being crossing entities, where two entities overlap but neither is contained in another. However, it is extremely rare, and there is only one occurrence of crossing entity in our datasets.}
Due to the tree structured representation used, the resulting algorithm has a time complexity that is {cubic} in $n$ for its inference procedure with $n$ being the number of words in the sentence.
This effectively makes the algorithm less scalable compared to models such as linear-chain CRF where the complexity is linear in $n$.
\citet{Lu:2015} proposed an alternative approach 
which shows a time complexity that is {linear} in $n$.
Their method differs from the conventional sequence labeling approach, in that a hypergraph representation was used in their model.

In this work, we make an observation that there exists an efficient model for recognizing overlapping mentions while still regarding the problem as a sequence labeling problem.
As opposed to the conventional approach where we assign labels to natural language words,
in our new approach we assign labels to the {\em gaps} between words, modeling the mention boundaries instead of modeling the role of words in forming mentions.
Furthermore, while these gap-based labels can be modeled using conventional graphical models like linear-chain CRFs, we also propose a novel multigraph representation to utilize such gap-based labels efficiently.
To the best of our knowledge, this is the first structured prediction model utilizing a gap-based annotation scheme to predict overlapping structures.

In this paper we make the following major contributions:
\squishlist
\item We propose a set of {\em mention separators} which can be collectively used to define all possible mention combinations together with a novel multigraph representation, on top of which efficient and exact inference can be performed.
\item Theoretically, we show that unlike a recently proposed state-of-the-art model that we compare against, our model does not exhibit the {\em spurious structures} issue in its learning procedure.
On the other hand, it still maintains the same inference time complexity as the previous model.
\item Empirically, we show that our model is able to achieve higher $F_1$-scores compared to previous models in multiple datasets.
\squishend

We believe our proposed approach and the novel representations can be applied in other research problems involving predicting overlapping structures, and we hope this work can inspire further research along such a direction. 

\vspace{-1mm}

\section{Related Work}

\vspace{-1mm}

NER or mention detection is normally regarded as a chunking task similar to base noun phrase chunking \cite{Kudo:2001,Shen:2005}, and hence the entities or mentions are usually represented in a similar way, using BILOU (\underline{B}eginning, \underline{I}nside, \underline{L}ast, \underline{O}utside, \underline{U}nit-length mention) or the simpler BIO
annotation scheme \cite{Ratinov:2009}. 
As a chunking task, it is commonly modeled using sequence labeling models, such as the linear-chain CRF \cite{Lafferty:2001}, which has time complexity $O(nT^2)$ with $n$ being the number of words in the sentence and $T$ the number of mention types.

On the task of recognizing mentions that may overlap with one another,
one of the earliest works that attempted to regard this task as a structured prediction task was by \citet{McDonald:2005}. They represented entity mentions as top-$k$ predictions with positive score from a structured multilabel classification model.
Their model has a time complexity of $O(n^3T)$.

\citet{Alex:2007} proposed a cascading approach using multiple linear-chain CRF models, each handling a subset of all the possible mention types, where the models which come later in the pipeline have access to the predictions of the models earlier in the pipeline.
This results in the time complexity of roughly $O(nT)$ depending on how the pipeline was designed.

\citet{Finkel:2009} later proposed a constituency parser to handle nested entities by converting each sentence into a tree, and each mention is represented as one of the subtrees. Their model has the standard time complexity for a constituency parser with binary grammar: $O(n^3\left\lvert{}G\right\rvert)$, where $\left\lvert{}G\right\rvert$ is the size of the grammar, which in this case is proportional to $T$ in the best case, and $T^3$ in the worst case. They showed that their model outperforms a semi-CRF baseline \cite{Sarawagi:2004} in terms of $F_1$-score.

Recently, \citet{Lu:2015} proposed a hypergraph-based model called {\it mention hypergraph} that is able to handle overlapping mentions with a linear time complexity $O(nT)$.
The model was shown to achieve competitive results compared to previous models on standard datasets.
As we will be making extensive comparisons against this previous state-of-the-art model, we will describe this approach in the next section.



\begin{figure}[t!]
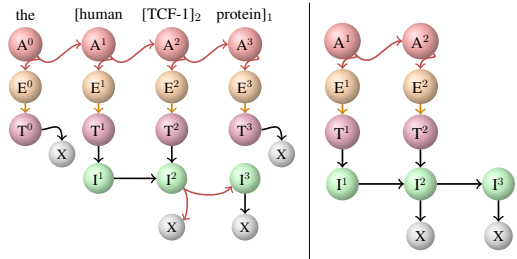

\centering
{\def\arraystretch{1.1}\tabcolsep=3pt
\setlength\doublerulesep{1pt}
\begin{tabular}{c|c}
\includestandalone[width=0.55\columnwidth]{anc/models-example1}\ &
\ \includestandalone[width=0.35\columnwidth]{anc/hypergraph-spurious-steps/hypergraph-spurious-4dd}
\end{tabular}
}
\vspace{-4mm}
\caption{(left) An example mention hypergraph encoding two overlapping mentions. (right) An example of spurious structure.}
\label{fig:hypergraph}
\vspace{-4mm}
\end{figure}

\section{Mention Hypergraph}\label{sec:mh}

In the mention hypergraph model of \citet{Lu:2015}, nodes and directed hyperedges\footnote{For brevity, in this paper we may also use {\em edge} to refer to {\em hyperedge} in some discussions.} are used together to encode mentions and their combinations.
The {\color{black}following five types} of {\color{black}nodes} are used at the position $k$ of a sentence:
\squishlist
\item $\mathbf{A}^k$ denotes all mentions starting at $k$ or later,
\item $\mathbf{E}^k$ denotes all mentions starting at $k$,
\item $\mathbf{T}^k_t$ denotes all mentions (type $t$) starting at $k$,
\item $\mathbf{I}^k_t$ denotes all mentions (type $t$) covering $k$,
\item $\mathbf{X}$ denotes the end of a mention (leaf node).
\squishend

Different hyperedges connecting these nodes are used to represent how the semantics of a node is composed from those of its child nodes.

Specifically, each $\mathbf{A}^k$ is connected to $\mathbf{A}^{k+1}$ and $\mathbf{E}^k$ through the hyperedge $\mathbf{A}^k\rightarrow(\mathbf{A}^{k+1},\mathbf{E}^k)$, denoting the fact that the set of mentions that start at $k$ or later is the union of the set of mentions that start at $k+1$ or later and the set of mentions that start at $k$. Each $\mathbf{E}^k$ is connected to $\mathbf{T}^k_1, \mathbf{T}^k_2, \ldots, \mathbf{T}^k_T$ through a hyperedge, denoting the fact that the mentions that start at $k$ must be one of the $T$ types. Each $\mathbf{T}^k_t$ can be connected to $\mathbf{I}^k_t$ through an edge (denoting there is a mention of type $t$ that starts at the $k$-th token) or to $\mathbf{X}$ through another edge (denoting there are no mentions of type $t$ that start at the $k$-th token). Each $\mathbf{I}^k_t$ can be connected to $\mathbf{I}^{k+1}_t$ (denoting there is a mention continuing to the next token), to $\mathbf{X}$ (denoting there is a mention ending here), or to both (with a single hyperedge, denoting the two cases above occur at the same time, a case of overlapping mentions).

In this mention hypergraph, each possible mention is represented as a path from a $\mathbf{T}$-node to the $\mathbf{X}$-node through a sequence of $\mathbf{I}$-nodes (each denoting the words which are part of the mention), and the set of all mentions present in a given sentence forms a hyperpath from the root node $\mathbf{A}_0$ to the leaf node $\mathbf{X}$.
Figure \ref{fig:hypergraph} shows how the mention hypergraph represents the two mentions in the phrase ``{\em the human TCF-1 protein}'', which are ``{\em TCF-1}'' and ``{\em human TCF-1 protein}''. The edges $\mathbf{T}^1-\mathbf{I}^1$ and $\mathbf{T}^2-\mathbf{I}^2$ respectively denote that the words ``{\em human}'' and ``{\em TCF-1}'' are the beginning of a mention, and the edges from the $\mathbf{I}$-nodes to the $\mathbf{X}$-node define the end of the mentions. We remark that any mention hypergraph which encodes the mentions in a sentence, like this example, forms a hyperpath from the root node $\mathbf{A}^0$ to the leaf node $\mathbf{X}$, where a hyperpath is defined as a subgraph of a hypergraph with the property that each node has exactly one outgoing (hyper)edge except the last node, and the root node is connected to all nodes.

We refer the readers to \citet{Lu:2015} for more details on the model.

\subsection{Spurious Structures}\label{sec:spurious}
Mention hypergraph is trained by maximizing the likelihood of the training data, similar to training a linear-chain CRF. Recall that the likelihood of the training data can be calculated by taking the score of the correct structures and divide it by the normalization term, which is the total score of all possible structures. \citet{Lu:2015} used a dynamic programming algorithm to calculate the normalization term. However, the normalization term calculated this way contains additional terms, which we call the {\em spurious structures}. This leads to the following:
\begin{theorem}
Let $Z'$ be the normalization term as calculated using forward-backward algorithm on mention hypergraph, and let $Z$ be the true normalization term. Then we have $Z' > Z$.
\end{theorem}

Due to space limitation, we provide a proof sketch here. We refer the reader to the supplemental material for the details on spurious structures.

\begin{proof}[Proof sketch]
First note that $Z'$ includes all possible hyperpaths, so $Z' \geq Z$. Next, due to the presence of a node with multiple parents (e.g., node $\mathbf{I}^2$ in Figure \ref{fig:hypergraph} (left)), $Z'$ includes the score of that node multiple times with different children, which results in a subgraph which is not a hyperpath. For example, $Z'$ includes the score\footnote{Note that structure scores $\exp(\mathbf{w}\cdot\mathbf{f})$ are always positive.} of the structure shown in Figure \ref{fig:hypergraph} (right), where node $\mathbf{I}^2$ has two children, and so it is not a hyperpath. Since $Z$ is the sum of all hyperpaths, this structure is not part of $Z$, but it is included in $Z'$, so $Z' > Z$.
\end{proof}

\vspace{-2mm}
Later we will see how this issue may affect the model's performance in predicting mentions.

\begin{figure}[t!]
\centering
\scalebox{1.0}
{
{\def\arraystretch{0.4}\tabcolsep=6pt
\setlength\doublerulesep{1pt}
\begin{tabular}{c|c|c|c}
$w\ \!\ \!\ \ \!w$ & $w\ \!\ {\color{red}[}w$ & $w{\color{red}]}\ \ \!w$ & $w{\color{red}]\ [}w$\\&&&\\
\texttt{\scriptsize{}X} & \texttt{\scriptsize{}S} & \texttt{\scriptsize{}E} & \texttt{\scriptsize{}ES}\\&&&\\\hline&&&\\
$w\ \!\ \!\text{\color{red}-}\ \!\ \!w$ & $w\ \!\ \!\color{red}\text{-}[\color{black}w$ & $w{\color{red}]\text{-}}\ \!\ \!w$ & $w{\color{red}]\text{-}[}w$\\&&&\\
\texttt{\scriptsize{}C} & \texttt{\scriptsize{}CS} & \texttt{\scriptsize{}EC} & \texttt{\scriptsize{}ECS}
\end{tabular}
}
}
\vspace{-2mm}
\caption{An illustration of the 8 mention separators. The opening bracket ({\color{red}[}), closing bracket ({\color{red}]}), and dash ({\color{red}-}) respectively refer to \texttt{S}, \texttt{E}, and \texttt{C}.}
\label{fig:separator}
\vspace{-3mm}
\end{figure}


\section{Mention Separators}\label{sec:mention-separators}

We now describe the mention separators which can be used to encode overlapping mentions in a sentence. Traditional encoding schemes that associate labels to words, such as BIO scheme, attach the semantics of the labels to the role of the words in forming mentions. For example, the label {\bf B} in BIO scheme denotes the role of the word it is attached to, which is the first word of a mention.

This BIO scheme cannot be used directly to encode overlapping mentions, since they only encode whether a word is part of a mention and possibly their position in the mention. We notice that by encoding the mention boundaries instead, we can represent overlapping mentions. This can be accomplished by assigning what we call {\em mention separators} to the gaps between two words.

\begin{figure*}[t!]
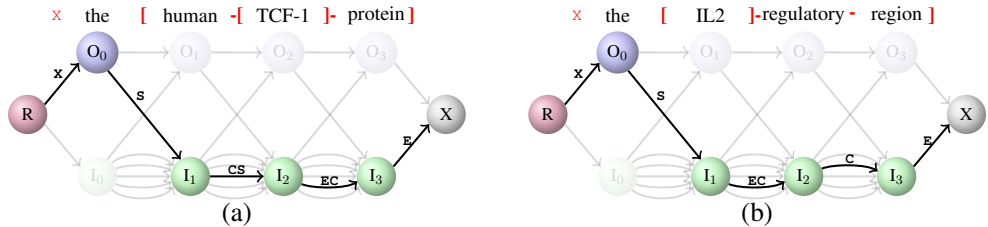

\centering
{\def\arraystretch{0}
\begin{tabular}{cc}
\includestandalone[width=0.9\columnwidth]{anc/models-example2} &
\includestandalone[width=0.9\columnwidth]{anc/models-example3} \\
(a) & (b)
\end{tabular}
}
\vspace{-3mm}
\caption{Our mention separator model with the \textsc{Edge} representation encoding two phrases.}
\label{fig:multigraph}
\vspace{-3mm}
\end{figure*}

At each gap, we consider eight possible types of mention separators based on the combination of the following three cases:
\squishenum
\item A mention is \underline{s}tarting at the next word (\texttt{S})
\item A mention is \underline{e}nding at the previous word (\texttt{E})
\item A mention is \underline{c}ontinuing to the next word (\texttt{C})
\squishend
Therefore, for each token, the possible combinations of cases are as follows: \texttt{ECS}, \texttt{EC}, \texttt{CS}, \texttt{C}, \texttt{ES}, \texttt{E}, \texttt{S}, and \texttt{X},  where \texttt{X} means none of the three cases applies.
For example, the separator \texttt{EC} means there is a mention ending at the current token and another mention (overlapping) continuing to the next token. Note that there might be more than just two mentions involved here.
Figure \ref{fig:separator} shows an illustration of these separators, and Figure \ref{fig:multigraph}a shows how they can be used to encode the example in Figure \ref{fig:hypergraph}.

Now we prove that the following theorem holds:
\begin{theorem}\label{thm:unique}
For any combination of mentions in a sentence, there is exactly one sequence of mention separators that encodes it.
\end{theorem}
\begin{proof}
Consider the gap between any two adjacent words in the sentence. The combination of mentions present in the sentence uniquely defines what mention separator is associated with this gap. If there is a mention starting at the next word, then case \texttt{S} applies. Similarly, if there is a mention ending at the previous word, case \texttt{E} applies. And finally, if there is a mention covering both words, case \texttt{C} applies. By combining the cases, we get the corresponding mention separator for this gap. In this way, each gap in the sentence has a unique mention separator, which in turn defines the unique sequence of mention separators.
\end{proof}

Note that the converse of Theorem \ref{thm:unique} is not true, as multiple mention combinations might encode to the same sequence of mention separators.

Now we describe two ways the mention separators can be used to encode overlapping mentions.

\vspace{-2mm}

\subparagraph{\textsc{State}-based} The first is by directly using these mention separators to replace the standard mention encoding scheme (e.g., BIO encoding) in standard linear-chain CRF. So we assign each mention separator to a state in a linear-chain CRF model. Since this model encodes the gap between words and also the gap before the first word and after the last word, a sentence with $n$ words is modeled by a sequence of $n+1$ mention separators. Since each sequence of mention separators can only encode mentions of the same type, we support multiple types by using multiple sequences, one for each mention type.

\vspace{-2mm}

\subparagraph{\textsc{Edge}-based} Now, we propose a novel way of utilizing these mention separators. Since the mention separators encode the gaps between words, it is more intuitive to assign the mention separators to the edges of a graphical model, as opposed to the states, as described in the previous paragraph. To do this, we need to define the states of the models in such a way that all possible sequences of mention separators are accounted for. For this purpose we assign two states to each word at position $k$:
\squishlist
\item $\mathbf{I}_k$: word at $k$ is part of a mention,
\item $\mathbf{O}_k$: word at $k$ is not part of any mentions.
\squishend

Next we define the edges between the states according to the eight possible mention separators between adjacent words.
More specifically, each mention separator is mapped to an edge connecting one state in the current position to another state in the next position depending on whether the separator defines current and next word as part of an mention, so in total we have eight edges between two positions in the model. 
Some mention separators may connect the same two states, for example, the \texttt{ES} and \texttt{C} separator both connect $\mathbf{I}_k$ to $\mathbf{I}_{k+1}$ since in both cases the current word and the next word are part of a mention. In those cases, we simply define multiple edges between the pair of states. The resulting graph, where there can be multiple edges between two states, is known in graph theory literature as a {\bf multigraph}\footnote{In this work, the  multigraph representation can also be regarded as a {\em lattice} where edges are associated with labels.}.

The first $\mathbf{I}$- and $\mathbf{O}$-nodes in the sentence are connected to the root node, and the last $\mathbf{I}$- and $\mathbf{O}$-nodes are connected to the unique leaf node $\mathbf{X}$.

Figure \ref{fig:multigraph}a shows how the \textsc{Edge}-based model encodes the two mentions ``{\it human TCF-1 protein}'' and ``{\it TCF-1}'' in the phrase ``{\em the human TCF-1 protein}'', and Figure \ref{fig:multigraph}b shows the encoding of the phrase found in the second example in Figure \ref{fig:main-example}.
Note how each edge maps to a distinct mention separator visualized in the text in red.

\begin{figure}[t!]
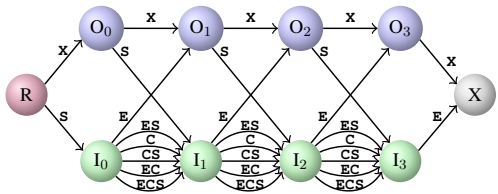

\centering
\includestandalone[width=0.9\columnwidth]{anc/multigraph-model}
\vspace{-3mm}
\caption{The full graph in \textsc{Edge}-based model.}
\label{fig:multigraph-model}
\vspace{-3mm}
\end{figure}

\begin{figure}[t]
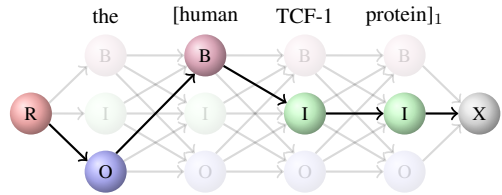

\centering
\includestandalone[width=0.9\columnwidth]{anc/linear-example}
\vspace{-3mm}
\caption{A linear-chain CRF model encoding a mention in BIO scheme.}
\vspace{-3mm}
\label{fig:linear-example}
\end{figure}

Figure \ref{fig:multigraph-model} shows the full graph of our \textsc{Edge}-based model, in a format similar to the trellis graph for linear-chain CRFs in Figure \ref{fig:linear-example}. We remark that the \textsc{Edge}-based model can be seen as an extension of linear-chain CRFs, with additional semantics attached to the edges.
Also note that this graph encodes only one mention type. To support multiple types, similar to the \textsc{State}-based approach we can use multiple chains, one for each type.

Note that the edges in our \textsc{Edge}-based representations are directed, with nodes on the left serving as parents to the nodes on the right.
Such directed edges will be helpful when performing inference, to be discussed in the next section.

We remark that the way we utilize multigraph in the \textsc{Edge}-based model can also be applied to the discontiguous mention model (DMM) by \citet{Muis:2016a}. In fact, it can be shown that the number of canonical structures as calculated in the supplementary material of DMM paper matches the number of possible paths in our multigraph-based model, as the transition matrix in DMM corresponds to the number of possible transitions from one position to the next position, which is encoded in our multigraph-based model as edges between adjacent positions. See the supplemental material for more discussion on this.

\subsection{Training, Inference and Decoding}


We follow the log-linear approach to define our model, using regularized log-likelihood in training data $\mathcal{D}$ as our objective function, as follows:

\vspace{-5mm}
{\small
\begin{equation}\label{eqn:obj}
\mathcal{L}_{\mathcal{D}}(\mathbf{w})\!=\!\!\sum_{(\mathbf{x},\mathbf{y})\in\mathcal{D}}\! \left[\ \sum_{\mathbf{e}\in\mathbf{y}}\mathbf{w}\cdot\mathbf{f}(\mathbf{e}) - \log Z_{\mathbf{w}}(\mathbf{x})\right]\! -\! \lambda\lvert\lvert\mathbf{w}\rvert\rvert^2
\end{equation}}

\vspace{-4mm}

Here, $(\mathbf{x},\mathbf{y})$ is a training instance consisting of the sentence $\mathbf{x}$ and the correct output $\mathbf{y}$, $\mathbf{w}$ is the weight vector, $\mathbf{f}(\mathbf{e})$ is the feature vector defined over the edge $\mathbf{e}$, $Z_{\mathbf{w}}(\mathbf{x})$ is the normalization term, and $\lambda$ is the $l_2$-regularization parameter. The objective function is then optimized until convergence using L-BFGS \cite{Liu:1989}.

\begin{table*}[ht!]
\centering
\footnotesize
{\def\arraystretch{1.1}\tabcolsep=2.2pt
\setlength\doublerulesep{1pt}
\begin{tabular}{l|rl|rl|rl|rl|rl|rl|rl|rl|rl}
& \multicolumn{6}{c|}{ACE-2004} & \multicolumn{6}{c|}{ACE-2005} & \multicolumn{6}{c}{GENIA}\\
& \multicolumn{2}{c|}{Train (\%)} & \multicolumn{2}{c|}{Dev (\%)} & \multicolumn{2}{c|}{Test (\%)} & \multicolumn{2}{c|}{Train (\%)} & \multicolumn{2}{c|}{Dev (\%)} & \multicolumn{2}{c|}{Test (\%)} & \multicolumn{2}{c|}{Train (\%)} & \multicolumn{2}{c|}{Dev (\%)} & \multicolumn{2}{c}{Test (\%)}
\\
\hline
\# sentence & 6,799 & & 829 & & 879 & & 7,336 & & 958 & & 1,047 & & 14,836 & & 1,855 & & 1,855 &\\
\ \  {\em w/ o.l.} & 2,685 & (39) & 293 & (35) & 373 & (42) & 2,686 & (37) & 341 & (36) & 330 & (32) & 3,199 & (22) & 366 & (20) & 448 & (24)\\
\# mentions & 22,207 & & 2,511 & & 3,031 & & 24,687 & & 3,217 & & 3,027 & & 46,473 & & 5,014 & & 5,600 &\\
\ \ {\em o.l.} & 10,170 & (46) & 1,091 & (43) & 1,418 & (47) & 9,937 & (40) & 1,192 & (37) & 1,184 & (39) & 8,337 & (18) & 915 & (18) & 1,217 & (22)\\
\ \ {\em o.l. (s)} & 5,431 & (24) & 624 & (25) & 780 & (26) & 5,044 & (20) & 600 & (19) & 638 & (21) & 4,613 & (10) & 479 & (10) & 634 & (11)\\
\end{tabular}
}
\vspace{-4mm}
\caption{Statistics of the datasets used in the experiments. {\em w/ o.l.}: sentences containing overlapping mentions; {\em o.l.}: overlapping mentions; {\em o.l. (s)}: overlapping mentions with the same type.}
\label{tbl:data-stat}
\vspace{-3mm}
\end{table*}

We note the mention hypergraph model also defines the objective in a similar manner.
For both of our models, the inference is done based on a generalized inside-outside algorithm.
Both models involve directed structures, on top of which the inference algorithm first
calculates the inside score for each node from the leaf node to root, and then the outside score
from the root to the leaf node, in very much the same way as 
how inference is done in a classic graphical model.
Specifically, for our \textsc{Edge}-based model, the inside scores are calculated using a bottom-up (right-to-left) dynamic programming procedure,
where we calculate the inside score at each node by summing up the scores associated with each path connecting the current node to one of its child nodes.
Each such path score is defined as the product of the inside score stored in that child node and the score defined over the edge connecting them.
The computation of the outside scores can be done in an analogous manner from left to right.
It can be verified that the time complexity of this inference procedure for our model is $O(nT)$, which is the same as the mention hypergraph model. Note that, however, both of our models do not have the spurious structures issue, as for any {\em path} in these models there are no nodes with multiple incoming edges.

During decoding, we perform MAP inference using a max-product procedure that is analogous to how the Viterbi decoding algorithm is used in conventional tree-structured graphical models to find out the highest-scoring subgraph, from which we extract mentions through the process that we call the {\em interpretation process}. As noted in previous section, there could be multiple mention combinations that correspond to the same sequence of mention separators, which presents an ambiguity during the interpretation process. For these ambiguous cases, we implemented the same interpretation process as that was done in the mention hypergraph model, which is by resolving ambiguous structures as nested mentions. For other cases, there is exactly one way to interpret the structure. For example, in Figure \ref{fig:multigraph}b, although there is only one gap marked as starting position (\texttt{S}) and two gaps marked as ending position (\texttt{EC} and \texttt{E}), the interpretation is clear that the two mentions here are ``{\em IL2}'' and ``{\em IL2 regulatory region}''.

\vspace{-2mm}

\section{Experiments}

\vspace{-1mm}

\subsection{Datasets}

\vspace{-1mm}

To assess our model's capability in recognizing overlapping mentions and make comparisons with previous models, we looked at datasets where overlapping mentions are explicitly annotated.
Following the previous work \cite{Lu:2015}, our main results are based on the standard ACE-2004 and ACE-2005 datasets \cite{Doddington:2004}.
We also additionally looked at the GENIA dataset \cite{Kim:2003}, which was used in the previous works \cite{Finkel:2009,Lu:2015}.

For ACE datasets, we used the same splits as used in our previous work \cite{Lu:2015}, published on our website\footnote{\url{http://statnlp.org/research/ie\#mention-hypergraph}}.
For GENIA, we used GENIAcorpus3.02p\footnote{\url{http://geniaproject.org/genia-corpus/pos-annotation}} that comes with POS tags for each word \cite{Tateisi:2004}.
Following previous works \cite{Finkel:2009,Lu:2015}, we first split the last 10\% of the data as the test set.
Next we used the first 80\% and the subsequent 10\% for training and development, respectively.
We made the same modifications as described by \citet{Finkel:2009} by collapsing all {\it DNA}, {\it RNA}, and {\it protein} subtypes into {\it DNA}, {\it RNA}, and {\it protein}, keeping {\it cell line} and {\it cell type}, and removing other mention types, resulting in 5 mention types.
The statistics of each {\color{black}dataset are shown} in Table~\ref{tbl:data-stat}.
We can see overlapping mentions are common in such datasets.

For more details on the dataset preprocessing, please refer to the supplemental material.

\begin{table*}[ht!]
\small
\centering
{\def\arraystretch{1.175}\tabcolsep=4.5pt
\setlength\doublerulesep{1pt}
\begin{tabular}{l||ccc|r|ccc|r||ccc|ccc}
& \multicolumn{4}{c|}{\multirow{2}{*}{ACE-2004}} & \multicolumn{4}{c||}{\multirow{2}{*}{ACE-2005}} & \multicolumn{3}{c|}{ACE-2004 } & \multicolumn{3}{c}{ACE-2005} \\
& \multicolumn{4}{c|}{\multirow{2}{*}{}} & \multicolumn{4}{c||}{} & \multicolumn{3}{c|}{($F_1$ optimized)} & \multicolumn{3}{c}{($F_1$ optimized)} \\
 & $P$ & $R$ & $F_1$ & $w/s$ & $P$ & $R$ & $F_1$ & $w/s$ &  $P$   &  $R$   &          $F_1$    &  $P$   &  $R$   &          $F_1$    \\
\hline
LCRF (single)   & 70.6 & 41.7 & 52.5 & 40.2 & 66.0 & 45.0 & 53.5 & 41.2 & 66.2 & 47.7 & 55.4 & 62.1 & 48.9 & 54.7  \\
LCRF (multiple) & 78.6 & 44.5 & 56.9 & 119.4 & 76.2 & 46.8 & 58.0 & 118.7 & 69.9 & 55.1 & 61.6 & 66.5 & 55.3 & 60.4  \\
\citet{Lu:2015}          & 81.2 & 45.9 & 58.6 & 472.5 & 78.6 & 46.9 & 58.7 & 516.6 & 72.5 & 55.7 &         63.0  & 66.3 & 57.3 &         61.5  \\
\hdashline
This work (\textsc{state}) & 78.0 & 51.2 & \textbf{61.8} & 50.5 & 75.3 & 51.7 & \textbf{61.3} & 52.1 & 71.2 & 58.0 & \textbf{64.0} & 67.6 & 58.4 & \textbf{62.7} \\
This work (\textsc{edge})  & 79.5 & 51.1 & \textbf{62.2} & 251.5 & 75.5 & 51.7 & \textbf{61.3} & 253.3 & 72.7 & 58.0 & \textbf{64.5} & 69.1 & 58.1 & \textbf{63.1} \\
\end{tabular}
}
\vspace{-2mm}
\caption{Main results (on ACE).}
\label{tab:ace}
\vspace{-4mm}
\end{table*}

\vspace{-1mm}

\subsection{Features}

For models that fall under the edge-based paradigm (mention hypergraph and our model), we define features over the edges in the models.
Features are defined as string concatenations of input features -- information extracted over the inputs (such as current word and POS tags of surrounding words) and output features -- structured information extracted over the output structure.
We carefully defined the input and output features in a way that {\color{black}allows us to make use} of the identical set of features for both our mention separator model and the baseline mention hypergraph model, in order to make a proper comparison.
We also followed \citet{Lu:2015} to add the additional  {\emph{mention penalty}} feature for our model and all baseline approaches so that we are able to tune $F_1$-scores on the development set.
Roughly speaking, the weight of this feature controls how confident the model should be in predicting more mentions. In other words, this is a way to balance the precision and recall of the model.

When defining the input features for both our model and the mention hypergraph model, we implemented the features used by previous works in each dataset based on the descriptions in their papers: we followed \citet{Lu:2015} for the features used in ACE datasets, and  \citet{Finkel:2009} for features used in GENIA dataset. 
In general, they include surrounding words, surrounding POS tags, bag-of-words, Brown clusters (for GENIA only), and orthographic features. 
See the supplemental material for more details.

\subsection{Experimental Setup}

We trained each model in the training set, then tuned the $l_2$-regularization parameter based on the development set.
For GENIA experiments, we also tuned the number of Brown clusters.
Following \citet{Lu:2015}, we also used each development set to tune the mention penalty to optimize the $F_1$-score and report the scores on the corresponding test sets separately.
Similar to \citet{Finkel:2009}, as another baseline model we also trained a standard linear-chain CRF using the BILOU scheme. Although this model does not support overlapping mentions, it gives us a baseline to see the extent to which our model's ability to recognize overlapping mentions can help the overall performance. There is also a simple extension\footnote{\color{edited}We also tried a more elaborate encoding scheme based on BIO scheme \citet{Tang:2013}, originally designed for discontiguous mentions, but is supposed to be able to also recognize overlapping mentions of the same type. However, the result is very similar to LCRF (multiple), perhaps due to the invalid structures issue noted by \citet{Muis:2016a}.} of this linear-chain CRF model that can support overlapping mentions of {\em different types} by considering each type separately using multiple chains, one for each type. We call this multiple-chain variant {\em LCRF (multiple)} and the earlier standard approach {\em LCRF (single)}.
In all models, we also implement the mention penalty feature, adapted accordingly so that increasing the feature weight will increase the number of mentions predicted by the model. See supplemental material for more details.

We implemented all models using Java, and also made additional comparisons on running time by running them under the same machine.
In addition, we also analyzed the convergence rate for different models.

\section{Results and Discussion}
\vspace{-2mm}



\subsection{Results on ACE}

Table \ref{tab:ace} shows the results on the ACE datasets, and these are our main results.
Following previous works \cite{Finkel:2009,Lu:2015}, we report standard precision ($P$), recall ($R$) and $F_1$-score percentage scores.
The highest results ($F_1$-score) and those results that are not significantly different from the highest results are highlighted in bold (based on bootstrap resampling test \cite{koehn2004statistical}, where $p>0.01$).
For ACE datasets, we make comparisons with the two versions of the linear-chain CRF baseline: LCRF (single) which does not support overlapping mentions at all and LCRF (multiple) which does not support overlapping mentions of the same type, as well as our implementation of the mention hypergraph  baseline \cite{Lu:2015}. 

From such empirical results we can see that our proposed model using mention separators consistently yields significantly better results ($p<0.01$) than the mention hypergraph model across these two datasets, under two setups (whether to optimize $F_1$-score or not).
Specifically, when the state-based approach is used (\textsc{State}), 
our approach is able to obtain a much higher recall, resulting in improved $F_1$-score.
Empirically, we found this approach was also faster than the LCRF baseline approach in terms of the number of words processed each second ($w/s$) during decoding, which is expected, since \textsc{State} uses fewer number of tags.\footnote{There are eight tags in \textsc{State} and nine in LCRF.}
The edge-based approach (\textsc{Edge}) using our proposed multigraph representation is able to achieve a significant speedup in comparison with the state-based approach. 
Although this model is still about 50\% slower than the mention hypergraph model\footnote{Though both models have the same time complexity, they differ by a constant factor.}, but it yielded a significantly higher $F_1$-score (up to 3.6 points higher on ACE-2004 before optimizing $F_1$-score).
These results largely confirm the effectiveness of our proposed mention separator model and the usefulness of the multigraph representation for learning the model.

And as expected, the LCRF baselines yields  relatively lower results compared to the other models, since it cannot predict overlapping mentions.\footnote{LCRF (single) cannot predict any overlapping mentions, while LCRF (multiple) cannot predict overlapping mentions of the same type.}
However, such results give us some idea on how much performance increase we can gain by properly recognizing overlapping mentions by looking at the results of LCRF (single), which in this case can be up to 9.7 points in $F_1$-score in ACE-2004.
We can also see the gain from recognizing overlapping mentions of the same type by looking at the results of LCRF (multiple), which can be up to 5.3 points in $F_1$-score in ACE-2004.

\begin{table}[t!]
\small
\centering
{\def\arraystretch{1.075}\tabcolsep=5.5pt
\setlength\doublerulesep{1pt}
\begin{tabular}{l|ccc|r}
 & $P$ & $R$ & $F_1$ & $w/s$\\
\hline
LCRF (single)       & 77.1 & 63.3 & 69.5 & 81.6\\
LCRF (multiple)     & 75.9 & 66.1 & \textbf{70.6} & 175.8 \\
\citet{Finkel:2009} & 75.4 & 65.9 & 70.3 & -\\
\citet{Lu:2015}     & 74.2 & 66.7 & 70.3 & 931.9\\
\hdashline
This work (\textsc{state}) & 74.0 & 67.7 & \textbf{70.7} & 110.8\\
This work (\textsc{edge}) & 75.4 & 66.8 & \textbf{70.8} & 389.2\\
\end{tabular}
}
\vspace{-4mm}
\caption{Results on GENIA.}
\label{tbl:genia-result-opt}
\vspace{-3mm}
\end{table}

\subsection{Results on GENIA}

Table \ref{tbl:genia-result-opt} shows the results of running the models with $F_1$-score tuning on GENIA dataset. All models include Brown clustering features learned from PubMed abstracts.
Besides the mention hypergraph baseline, we also make comparisons with the system of \citet{Finkel:2009} that can also support overlapping mentions.

We see that the mention hypergraph model matches the performance of the constituency parser-based model of \citet{Finkel:2009}, while our models based on mention separators yield significantly higher scores (\mbox{$p<0.05$}) than all other baselines (except LCRF (multiple), which we will discuss shortly). There are two observations worth mentioning: (1) the absolute difference of $F_1$-scores of our models and the baseline models in GENIA is much smaller compared to that in ACE datasets, and (2) the LCRF (multiple) model in GENIA dataset can achieve higher scores compared to other more complex baseline models, although LCRF (multiple) does not support overlapping mentions of the same type. We suspect that these two observations are due to the small proportion of overlapping mentions in GENIA (18\%, as compared to >40\% in ACE datasets, see Table \ref{tbl:data-stat}). To investigate this, we conduct a few more sets of experiments.

\subsection{Further Experiments}

\paragraph{On different types of sentences:} 
As these datasets consist of both overlapping and non-overlapping mentions,
to further understand the model's effectiveness in recognizing overlapping mentions (and non-overlapping mentions), we performed some additional experiments on the mention hypergraph model and our model.\footnote{We also performed this on other models. Due to space constraint, we do not include {\color{edited}the results} here. See the supplemental material for more details.}
Specifically, we split the test data into two portions, one that consists of only sentences that contain overlapping mentions (O) and those which do not (\O).
The results are shown in Table \ref{tbl:result-overlapping}.

\begin{table}[t!]
\small
\centering
{\def\arraystretch{1.075}\tabcolsep=3.6pt
\setlength\doublerulesep{1pt}
\begin{tabular}{lc|c|ccc|ccc}
& & \multirow{2}{*}{\%} & \multicolumn{3}{c|}{{\scriptsize{\citet{Lu:2015}}}}  & \multicolumn{3}{c}{\scriptsize{This work (\textsc{Edge})}} \\
& & & \multicolumn{1}{c}{$P$} & \multicolumn{1}{c}{$R$} & \multicolumn{1}{c|}{$F_1$} & \multicolumn{1}{c}{$P$} & \multicolumn{1}{c}{$R$} & \multicolumn{1}{c}{$F_1$}\\
\hline
\multirow{2}{*}{ACE-2004} & O & 42 & 72.5 & 52.4 & 60.8 & 72.1 & 55.3 & 62.6 \\
& \O & 58 & 72.5 & 65.0 & 68.6 & 74.1 & 65.5 & 69.5 \\\hline
\multirow{2}{*}{ACE-2005} & O & 32 & 68.1 & 52.6 & 59.4 & 70.4 & 55.0 & 61.8 \\
& \O & 68 & 64.1 & 65.1 & 64.6 & 67.2 & 63.4 & 65.2 \\\hline
\multirow{2}{*}{GENIA} & O & 24 & 76.3 & 60.8 & 67.7 & 76.5 & 60.3 & 67.4 \\
& \O & 76 & 73.1 & 70.7 & 71.9 & 74.8 & 71.3 & 73.0\\
\end{tabular}
}
\vspace{-2mm}
\caption{Results on different types of sentences.}
\label{tbl:result-overlapping}
\vspace{-3mm}
\end{table}

\begin{figure*}[ht!]
\begin{center}
\scalebox{0.95}
{
\includegraphics[trim=10mm 4.5mm 2mm 2.5mm,width=0.33\linewidth]{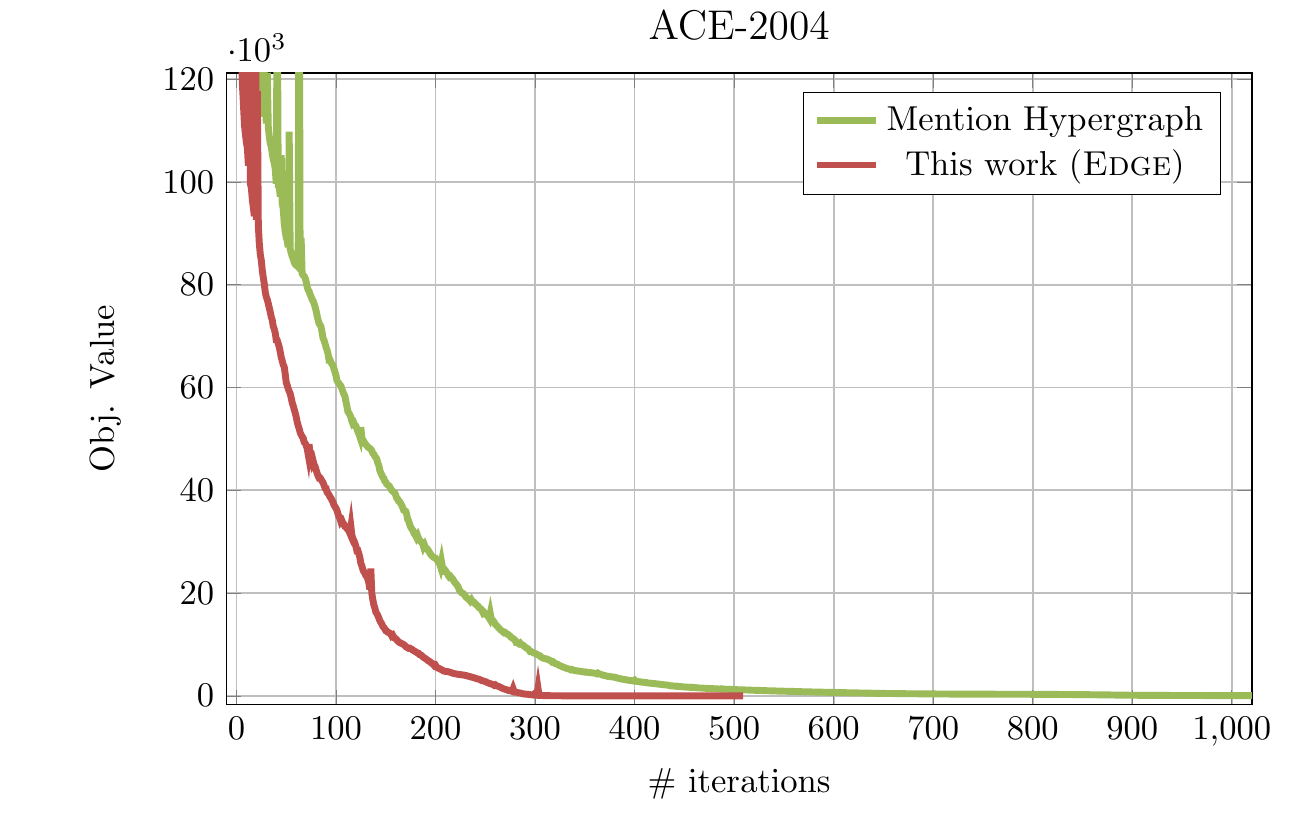}~
\includegraphics[trim=10mm 4.5mm 2mm 2.5mm,width=0.33\linewidth]{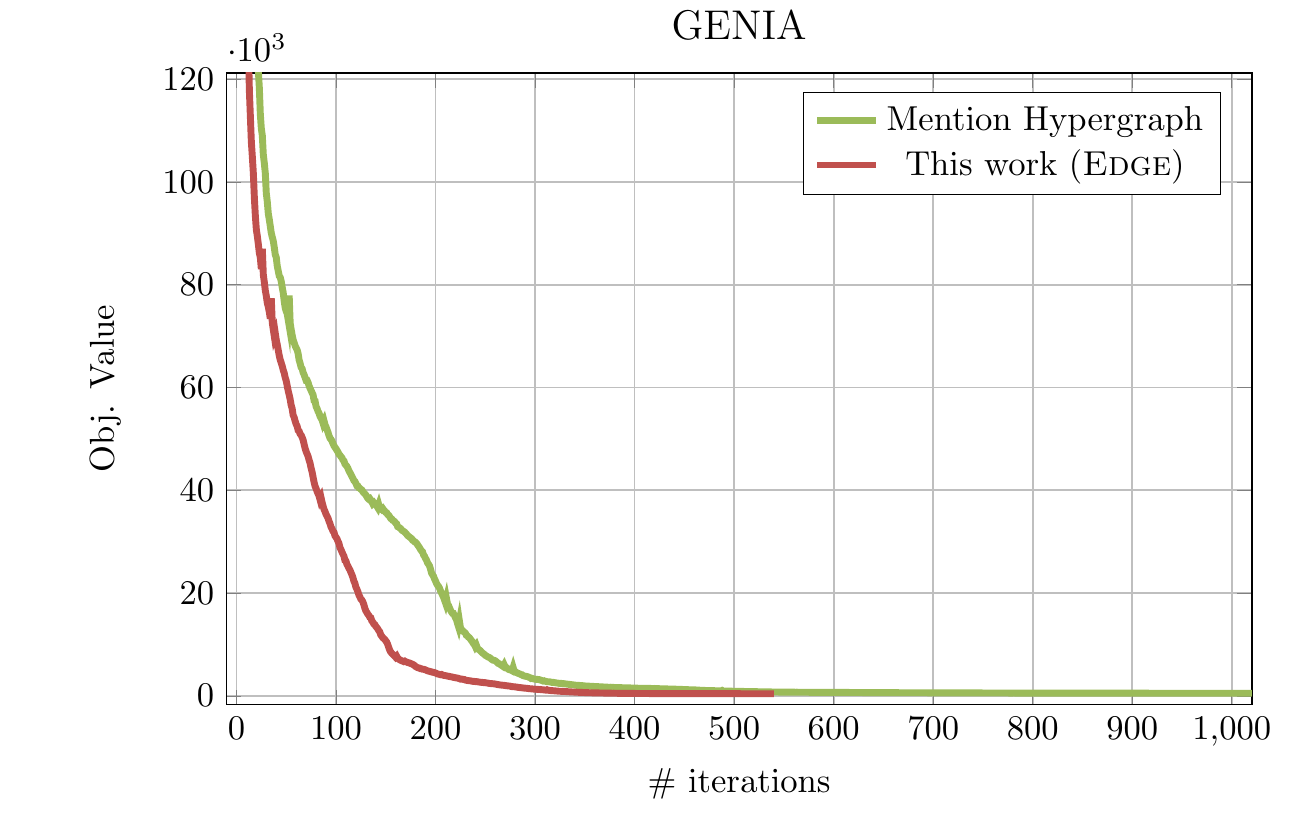}~
\includegraphics[trim=10mm 4.5mm 2mm 2.5mm,width=0.33\linewidth]{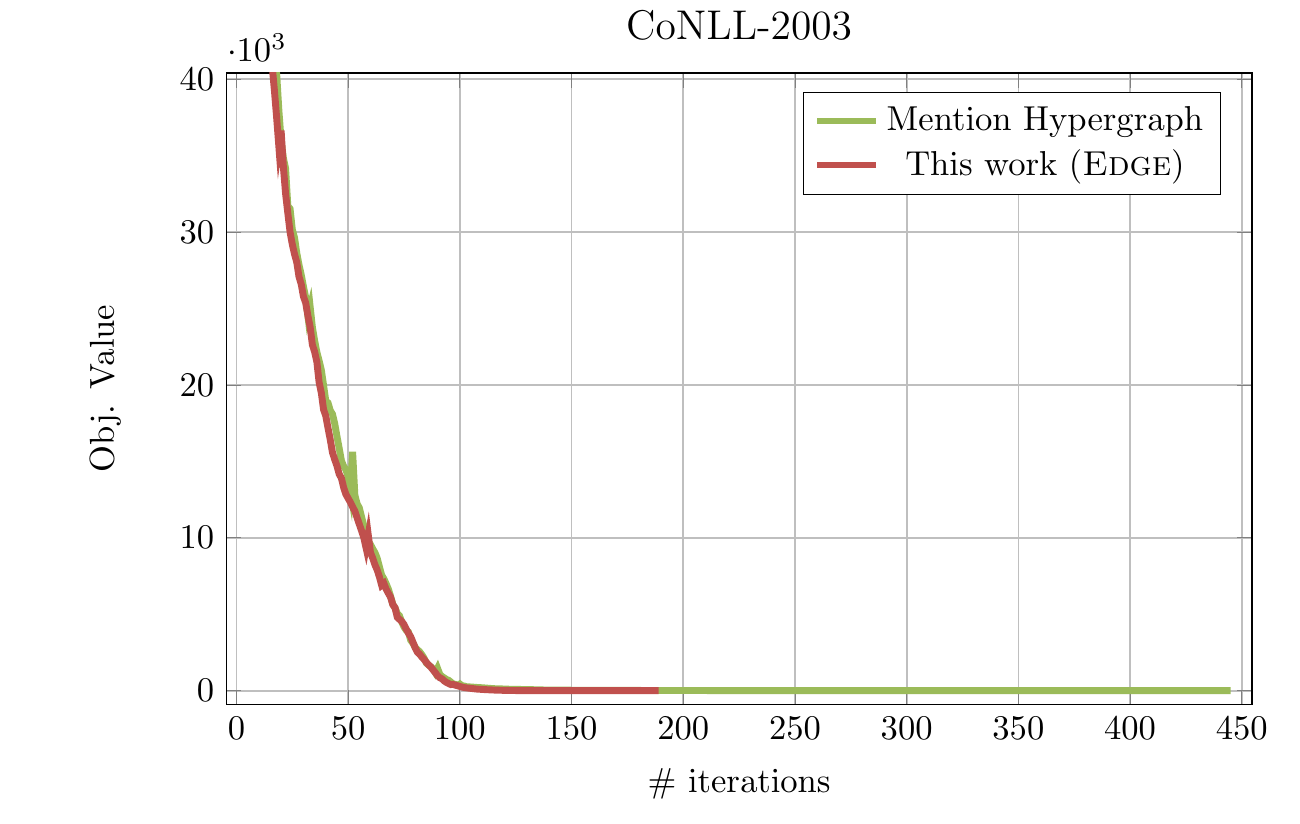}
}
\end{center}
\vspace{-5mm}
\caption{Objective vs. training iterations.}
\label{fig:obj-value}
\vspace{-4mm}
\end{figure*}

We can see that in ACE datasets, our model achieves  higher $F_1$-scores compared to the mention hypergraph for both portions, but it achieves slightly lower results in GENIA dataset for the portion that contains overlapping mentions. 
We believe that our models learn parameters so as to obtain an optimal overall performance, and since the proportion of the overlapping mentions in GENIA is much smaller compared to that in ACE datasets, it learns to focus more on the non-overlapping mentions. This is supported by the fact that the difference of $F_1$-score between the mention hypergraph model and our model in GENIA is larger compared to the difference in ACE datasets (1.1 points in GENIA, compared to 0.9 and 0.6 points in ACE).

These results also lead to the interesting empirical finding that our model appears to be able to do well also on recognizing non-overlapping mentions. 
This motivates us to conduct the next set of experiments.

\vspace{-2mm}

\begin{table}[t!]
\small
\centering
{\def\arraystretch{1.075}\tabcolsep=4.78pt
\setlength\doublerulesep{1pt}
\begin{tabular}{l|ccc|r}
 & $P$ & $R$ & $F_1$ & $w/s$\\\hline
LCRF (single)   & 84.2 & 83.5 & \textbf{83.8}  & 148.6\\
LCRF (multiple) & 91.5 & 78.2 & \textbf{84.3}  & 283.4\\
\citet{Ratinov:2009}&-&-&83.7&-\\
\citet{Lu:2015}          & 91.1 & 77.0 & 83.5  & 1169.7\\
\hdashline
This work (\textsc{state}) & 91.1 & 78.2 & \textbf{84.2}  & 116.3\\
This work (\textsc{edge})  & 91.3 & 78.2 & \textbf{84.3}  & 554.0\\
\end{tabular}
}
\vspace{-2mm}
\caption{Results on CoNLL-2003 (without optimizing $F_1$-score).}
\label{tab:conll}
\vspace{-4mm}
\end{table}

\paragraph{On data without overlapping mentions:} We also performed  one additional set of experiments, on the standard CoNLL-2003 dataset \cite{tjong2003introduction}, which has no overlapping mentions.

The results (without optimizing $F_1$-score) are shown in Table \ref{tab:conll}.
We see that our models based on mention separators outperform  baseline models such as the Illinois NER system where external resources are not used \cite{Ratinov:2009}, and a linear-chain CRF model, although the linear-chain CRF baseline models some interactions between distinct mention types and our models do not.
Such results also suggest that modeling the interactions between distinct mention types may not be crucial to get a good performance in mention recognition. This is further corroborated by the result of LCRF (multiple), which is higher than the result of LCRF (single) by about 0.5 points.

When comparing our model against the mention hypergraph model,
we note that our model consistently yields a higher recall.
We speculate  this is due to the fact that as our model does not exhibit the issue of spurious structures we discussed in Section \ref{sec:spurious}, it is more confident in making its predictions.

\paragraph{On convergence:}

We also empirically analyzed the convergence properties of the two models.
Empirically, as illustrated in Figure \ref{fig:obj-value} which shows how the objective improves when the training progresses on ACE-2004, GENIA, and CoNLL-2003, we found that {\color{edited}our \textsc{Edge}-based model} requires significantly less iterations to converge than the mention hypergraph on the former two datasets which contain overlapping mentions. 
We believe it is possible that this slower convergence is due to the spurious structures issue in mention hypergraphs, which causes the objective function to be more complex to optimize.
However, some further analyses on the convergence issue and the impact of different ways of exploiting features (over different hyperedges) for the hypergraph-based models are needed.

\vspace{-1mm}

\section{Conclusion and Future Work}

\vspace{-1mm}

We proposed the novel {\em mention separators} for mention recognition where mentions may overlap with one another. We also proposed two ways these mention separators can be utilized to encode overlapping mentions, where one of them utilizes a novel multigraph-based representation.
We showed that by utilizing mention separators, we can get better recognition results compared to previous models, and by utilizing the multigraph representation, we can maintain a good inference speed, {\color{edited}albeit still slower than the mention hypergraph model.}
We also performed theoretical analysis on the model and showed that our model does not present the {\em spurious structures} issue associated with a previous state-of-the-art model, while still keeping the same inference time complexity.

Future work includes further investigations on how to apply the multigraph approach to other structured prediction tasks, as well as applications of the proposed model in other related NLP tasks that involve the prediction of overlapping structures, such as equation parsing \cite{Roy:2016}.

The code used in this paper is available at \mbox{\url{http://statnlp.org/research/ie/}}.

\vspace{-1mm}

\section*{Acknowledgments}

\vspace{-1mm}

We thank all the reviewers for their useful feedback to the earlier draft of this paper. This work is supported by MOE Tier 1 grant SUTDT12015008.

\bibliography{emnlp2017-mention-separators}
\bibliographystyle{emnlp_natbib}
\vphantom{.}
\end{document}


\maketitle
\begin{abstract}
This is the supplementary material for ``Labeling Gaps Between Words: Recognizing Overlapping Mentions with Mention Separators'' \cite{Muis2017b}. This material explains in more depth the issue of spurious structures and also the experiments settings.
\end{abstract}

\section{Details on Spurious Structures}
About mention hypergraph, we remarked in Section 3.1 that the normalization term calculated by the forward-backward algorithm includes spurious structures, which are structures that are not part of the true normalization term. This section shows in more details how this is the case using some examples.

\begin{figure}[ht!]
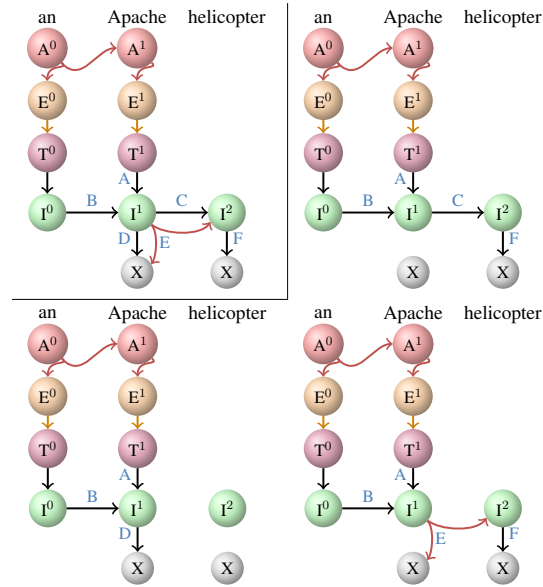

\centering
\begin{tabular}{c|c}
\includestandalone[width=0.45\columnwidth]{anc/hypergraph-spurious} &
\includestandalone[width=0.45\columnwidth]{anc/hypergraph-spurious-a} \\\cline{1-1}
\multicolumn{1}{c}{\includestandalone[width=0.45\columnwidth]{anc/hypergraph-spurious-b}} &
\includestandalone[width=0.45\columnwidth]{anc/hypergraph-spurious-c} \\
\end{tabular}
\vspace{-4mm}
\caption{(top left) A simplified example of a mention hypergraph with restricted edges. (others) The three possible (hyper-)paths from the root node to the leaf node.}
\label{fig:hypergraph-spurious}
\vspace{-4mm}
\end{figure}

Consider the simplified mention hypergraph as shown in Figure \ref{fig:hypergraph-spurious} (top left) consisting of three words and where the possible edges have been restricted to what are shown in the figure. Also, let A, B, C, D, E, F respectively denote the edges $\mathbf{T}^1\rightarrow(\mathbf{I}^1)$, $\mathbf{I}^0\rightarrow(\mathbf{I}^1)$, $\mathbf{I}^1\rightarrow(\mathbf{I}^2)$, $\mathbf{I}^1\rightarrow(\mathbf{X})$, $\mathbf{I}^1\rightarrow(\mathbf{I}^2,\mathbf{X})$, and $\mathbf{I}^2\rightarrow(\mathbf{X})$ as shown in Figure \ref{fig:hypergraph-spurious} (left). Further assume that features are only defined on these labeled edges.

Recall that in graphical models, any prediction by the model forms a {\it (hyper-)path} from the root node (here $\mathbf{A}^0$) to the leaf node ($\mathbf{X}$), which means each node other than the leaf node has exactly one outgoing (hyper-)edge. Now, notice that there are only three possible paths here, one for each of the three (hyper-)edges coming out from the node $\mathbf{I}^1$ associated with the word ``Apache''. See the top right, bottom left, and bottom right of Figure \ref{fig:hypergraph-spurious} for the visualization.

\begin{figure*}[h!]
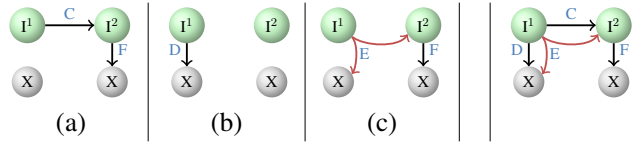

\centering
\begin{tabular}{c|c|c|c|c}
\includestandalone{anc/hypergraph-spurious-steps/hypergraph-spurious-1a} &
\includestandalone{anc/hypergraph-spurious-steps/hypergraph-spurious-1b} &
\includestandalone{anc/hypergraph-spurious-steps/hypergraph-spurious-1c} & \quad &
\includestandalone{anc/hypergraph-spurious-steps/hypergraph-spurious-1}\\
(a) & (b) & (c) &&
\end{tabular}
\caption{(left) The graphical representation of all 3 mention combinations represented by the node $\mathbf{I}_1$. (right) The full graph rooted at $\mathbf{I}^1$.}
\label{fig:hyp-spur-1}
\vspace{-2mm}
\end{figure*}

Now recall that in mention hypergraph, each node is assigned a certain set of mention combinations which the node represents, as defined by the subgraph from this node to the leaf node $\mathbf{X}$. In particular, the node $\mathbf{I}^1$ encodes three partial mentions (partial because the start of the mentions are undefined yet, as they are defined by the edge from $\mathbf{T}$-nodes to $\mathbf{I}$-nodes): \{``Apache helicopter''\}, \{``Apache''\}, and \{``Apache'',``Apache helicopter''\}, as shown in Figure \ref{fig:hyp-spur-1}. Similarly, we can see the subgraphs rooted at nodes $\mathbf{A}^1$ and $\mathbf{E}^0$ in Figure \ref{fig:hyp-spur-2} and \ref{fig:hyp-spur-3}, respectively.

\begin{figure*}[h!]
\centering
\begin{tabular}{c|c|c|c|c}
\includestandalone{anc/hypergraph-spurious-steps/hypergraph-spurious-2a} &
\includestandalone{anc/hypergraph-spurious-steps/hypergraph-spurious-2b} &
\includestandalone{anc/hypergraph-spurious-steps/hypergraph-spurious-2c} & \quad &
\includestandalone{anc/hypergraph-spurious-steps/hypergraph-spurious-2}\\
(a) & (b) & (c) &&
\end{tabular}
\caption{(left) The graphical representation of all 3 mention combinations represented by the node $\mathbf{A}_1$. (right) The full graph rooted at $\mathbf{A}^1$.}
\label{fig:hyp-spur-2}
\vspace{-2mm}
\end{figure*}

\begin{figure*}[h!]
\centering
\begin{tabular}{c|c|c|c|c}
\includestandalone{anc/hypergraph-spurious-steps/hypergraph-spurious-3a} &
\includestandalone{anc/hypergraph-spurious-steps/hypergraph-spurious-3b} &
\includestandalone{anc/hypergraph-spurious-steps/hypergraph-spurious-3c} & \quad &
\includestandalone{anc/hypergraph-spurious-steps/hypergraph-spurious-3}\\
(a) & (b) & (c) &&
\end{tabular}
\caption{(left) The graphical representation of all 3 mention combinations represented by the node $\mathbf{E}_0$. (right) The full graph rooted at $\mathbf{E}^0$.}
\label{fig:hyp-spur-3}
\vspace{-2mm}
\end{figure*}

\begin{figure*}[h!]
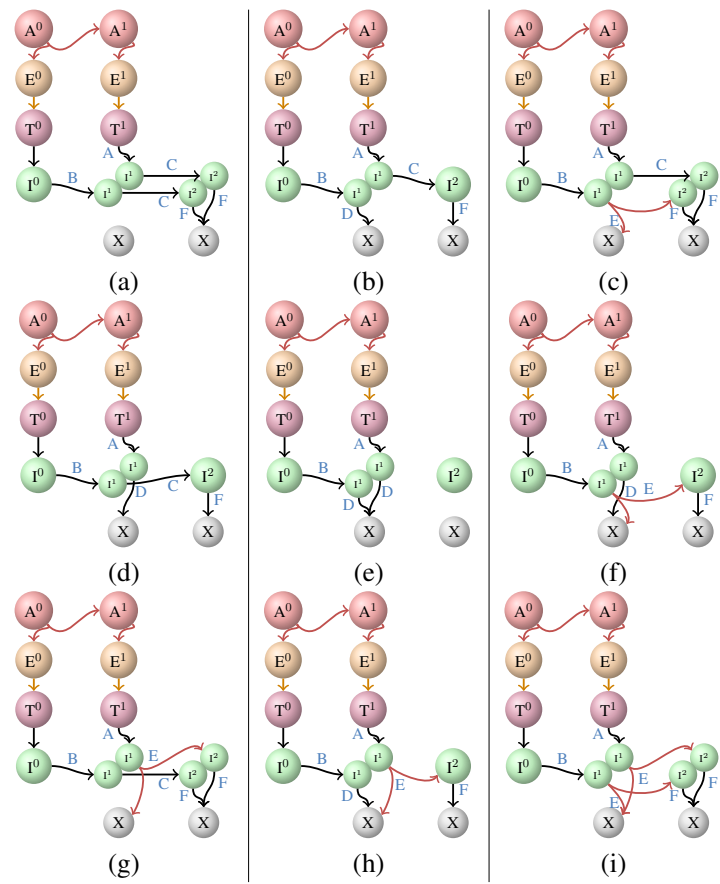

\centering
\begin{tabular}{c|c|c}
\includestandalone{anc/hypergraph-spurious-steps/hypergraph-spurious-4a} &
\includestandalone{anc/hypergraph-spurious-steps/hypergraph-spurious-4b} &
\includestandalone{anc/hypergraph-spurious-steps/hypergraph-spurious-4c} \\
(a) & (b) & (c) \\
\includestandalone{anc/hypergraph-spurious-steps/hypergraph-spurious-4d} &
\includestandalone{anc/hypergraph-spurious-steps/hypergraph-spurious-4e} &
\includestandalone{anc/hypergraph-spurious-steps/hypergraph-spurious-4f} \\
(d) & (e) & (f) \\
\includestandalone{anc/hypergraph-spurious-steps/hypergraph-spurious-4g} &
\includestandalone{anc/hypergraph-spurious-steps/hypergraph-spurious-4h} &
\includestandalone{anc/hypergraph-spurious-steps/hypergraph-spurious-4i} \\
(g) & (h) & (i)
\end{tabular}
\caption{The graphical representation of all 9 mention combinations represented by node $\mathbf{A}^0$. See main text for details.}
\label{fig:hyp-spur-4}
\vspace{-4mm}
\end{figure*}

Now, the node $\mathbf{A}^0$ includes the 9 possible entity combinations which are the results of taking all possible combinations in $\mathbf{A}^1$ and $\mathbf{E}^0$. The list of mention combinations and the respective paths used to represent each of these combinations is as follows:
\squishletter
\item A-C-F and B-C-F ({\it Apache helicopter}, {\it an Apache helicopter})
\item A-C-F and B-D ({\it Apache helicopter}, {\it an Apache})
\item A-C-F and B-E-F ({\it Apache helicopter}, {\it an Apache helicopter}, {\it an Apache})
\item A-D and B-C-F ({\it Apache}, {\it an Apache helicopter})
\item A-D and B-D ({\it Apache}, {\it an Apache})
\item A-D and B-E-F ({\it Apache}, {\it an Apache helicopter}, {\it an Apache})
\item A-E-F and B-C-F ({\it Apache helicopter}, {\it Apache}, {\it an Apache helicopter})
\item A-E-F and B-D ({\it Apache helicopter}, {\it Apache}, {\it an Apache})
\item A-E-F and B-E-F ({\it Apache helicopter}, {\it Apache}, {\it an Apache helicopter}, {\it an Apache})
\squishend
See Figure \ref{fig:hyp-spur-4} for a graphical representation of those 9 mention combinations. We display some nodes twice to highlight the different paths representing each distinct mention, a result of considering the mentions in $\mathbf{A}^1$ and $\mathbf{E}^0$ independently. Notice that this differing paths exist because the node $\mathbf{I}^1$ has two incoming edges in the path from $\mathbf{A}^0$ to $\mathbf{I}^1$. Further note that these are only the graphical representations of the mention combinations represented by the root node, as defined by the scoring of the structures in the objective function.

The crucial observation is that out of those nine mention combinations, only three of them form valid paths of the original graph in Figure \ref{fig:hypergraph-spurious}, namely: (a), (e), and (i). The other six mention combinations will never be predicted by the model, as they do not form paths found in the full graph. For example, in (b), the node $\mathbf{I}^1$ has two outgoing edges: C and D, and so this graph does not form a path.

Those six mention combinations which are part of the structures represented by the root node $\mathbf{A}^0$ which do not form valid paths are the spurious structures, as during training they are calculated as part of the normalization term, but the model can not output any of those as a prediction, since they do not form a path.

In essence, the normalization term fails to take into account the restriction of forming a path when calculating the scores of all possible paths because it is considering the two sub-paths from $\mathbf{A}^1$ and $\mathbf{E}^0$ independently, which cannot capture the restriction that the node $\mathbf{I}^1$ should have only one outgoing edge. This is what causing the spurious structures issue in mention hypergraph.

As a final remark, note that, depending on the heuristics used, when the model outputs structure (i) as its prediction, we can still interpret that structure to mean any of the six mention combinations. More technically, this means that the spurious structures do not affect the interpretation process, but they do affect how the objective function is calculated, which in turns affects how the learning process goes. And as can be seen from the experiments, removing these spurious structures from the objective function indeed improves the entity recognition ability of the model.

\section{Mention Separators}
This section will give more examples and illustrations on how mention separators can be used to encode any overlapping mentions in a sentence, following the description at Section 4 in the main paper.

\begin{figure*}[ht!]
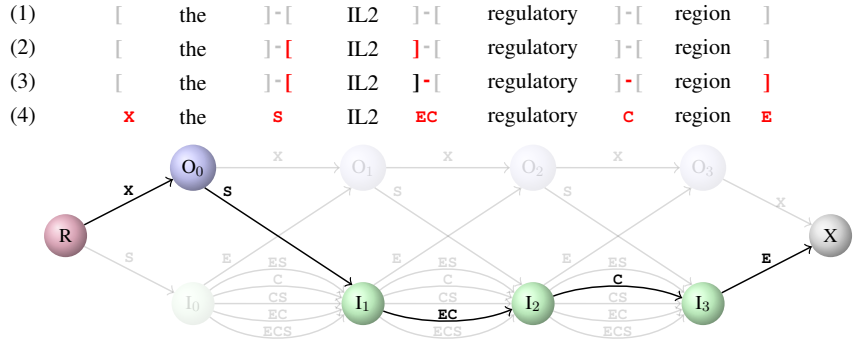

\centering
\includestandalone[width=0.75\textwidth]{anc/multigraph-example}
\vspace{-2mm}
\caption{Example on how to encode mentions using mention separators with the \textsc{Edge}-based model.}
\label{fig:multigraph-example}
\vspace{-2mm}
\end{figure*}

Suppose we want to encode the phrase ``the IL2 regulatory region'' containing two DNA mentions: ``IL2'' and ``regulatory region'' (example taken from GENIA dataset). In Figure \ref{fig:multigraph-example}, we start in step (1) by taking note of the possible starting, continuing, or ending marker in the gaps between the words. Then in step (2) we process the mention ``IL2'', marking the starting and ending marker before and after the mention accordingly. In step (3) we process the mention ``IL2 regulatory region'', marking the starting, continuing, and ending marker accordingly. Note that here it shares the starting marker as the previous mention. After all mentions have been processed, at step (4) we convert the combination of markers at each gap to the corresponding mention separator. Finally, we model the resulting sequence of mention separators into the \textsc{Edge}-based model, selecting the edge corresponding to the mention separators.

The decoding process will be the reverse of the encoding process, while using some heuristics to interpret the mention separator sequence. Continuing the example, given the path in the \textsc{Edge}-based model, we extract the sequence of mention separators, and then convert them into the markers at each gap, resulting in the one we see at step (3). Now, we need to interpret this sequence to find out what are the mentions encoded by this sequence. First we note that since the end marker after the word ``region'' is active, that means there is a mention ending at that word. And since the only active starting marker is the one before the word ``IL2'', it must be the case that ``IL2 regulatory region'' is {\it one} of the mentions encoded by this sequence. This mention already explains the presence of most active markers, except the end marker after the word ``IL2''. This means there is another mention ending with the word ``IL2''. And again, since there is only one active starting marker, we conclude that ``IL2'' is another mention encoded by the sequence. Finally, we note that these two mentions already explain the mention separator sequence (i.e., these two mentions, when encoded, will result in the same mention separator sequence that we have), and so we end the interpretation process.

Note that the example in Figure \ref{fig:multigraph-example} shows only the graph for recognizing one type. In the full model, there will be multiple chains, one for each type. An example of the full model can be seen in Figure \ref{fig:multigraph-full-example}.

\begin{figure*}[ht!]
\centering
\includestandalone[width=\textwidth]{anc/multigraph-full-example}
\vspace{-8mm}
\caption{The multigraph model with two chains representing two types: DNA and PROT.}
\label{fig:multigraph-full-example}
\end{figure*}

\subsection{Relation to Previous Work}
We also want to remark that the number of paths in our multigraph-based model can be calculated in the same way as how we previously calculated the number of canonical encoding for discontiguous mention recognition model, which is based on the mention hypergraph, using a transition matrix \cite{Muis:2016a}. In fact, the number of edges between two states in our multigraph-based model reflects the numbers in the transition matrix, and so this multigraph-based model can be seen as a model that utilizes the canonical structures found in the mention hypergraph model. This means the way we use multigraph to represent the overlapping mentions as modeled by the mention hypergraph model theoretically can also be applied to the discontiguous mention model.

However, the number of edges in the multigraph representation will explode. To illustrate, in mention hypergraph, since there is only 1 node per word (excluding the $\mathbf{A}$, $\mathbf{E}$, and $\mathbf{T}$ nodes), the multigraph-based model requires only $2^1 = 2$ states with $2^3 = 8$ edges per word, while in the discontiguous mention model supporting three components, since there are 5 nodes per word ($\mathbf{B}_0$, $\mathbf{O}_1$, $\mathbf{B}_1$, $\mathbf{O}_2$, and $\mathbf{B}_2$), the multigraph-based model will require $2^5 = 32$ nodes, with average number of edges per word being $2^{13} = 8192$, which makes it much slower than the hypergraph-based counterpart with only $25$ edges per word. So we can say that our multigraph-based approach trades off the speed in mention hypergraph with a higher $F_1$-score.

\section{Features}

For ACE datasets we used these features:
\squishenum
\item Words and POS tags (with window of 3 words to the left and right of current word)
\item Words and POS tags n-gram (up to length 4 containing current word)
\item Bag-of-word (with window of 5 words to the left and to the right of current word)
\item Orthographic (following \citet{Lu:2015})
\item Parent node type
\squishend

\begin{table*}[ht!]
\small
\centering
{\def\arraystretch{1.1}\tabcolsep=4.78pt
\setlength\doublerulesep{1pt}
\begin{tabular}{l||ccc|ccc||ccc|ccc||r}
& \multicolumn{3}{c|}{\multirow{2}{*}{CoNLL2003-dev}} & \multicolumn{3}{c||}{\multirow{2}{*}{CoNLL2003-test}} & \multicolumn{3}{c|}{CoNLL2003-dev} & \multicolumn{3}{c||}{CoNLL2003-test}&\multirow{3}{*}{$w/s$}\\
& \multicolumn{3}{c|}{} & \multicolumn{3}{c||}{} & \multicolumn{3}{c|}{($F$ optimized)} & \multicolumn{3}{c||}{($F$ optimized)} \\
 & $P$ & $R$ & $F$ & $P$ & $R$ & $F$  & $P$ & $R$ & $F$ & $P$ & $R$ & $F$ & \\\hline
LCRF (single) & 90.0 & 88.9 & \textbf{89.5} & 84.2 & 83.6 & \textbf{83.9}  & 90.1 & 88.9 & \textbf{89.5} & 84.3 &  83.4 & 83.8 & 148.6\\
LCRF (multiple) & 94.4 & 84.6 & \textbf{89.2} & 91.5 & 78.2 & \textbf{84.3}  & 92.7 & 86.8 & \textbf{89.6} & 88.6 &  81.2 & \textbf{84.7} & 283.4\\
\citet{Ratinov:2009}&-&-&89.3&-&-&83.7&-&-&-&-&-&-&-\\
\citet{Lu:2015}          & 94.4 & 83.4 & 88.5 & 91.1 & 77.0 & 83.5  & 89.7 & 88.7 & 89.2 & 84.6 & 82.9 & 83.8 & 1169.7\\
\hdashline
This work (\textsc{state}) & 94.2 & 84.7 & \textbf{89.2} & 91.1 & 78.2 & \textbf{84.2}  & 91.2 & 88.1 & \textbf{89.6} & 86.3 & 82.4 & \textbf{84.3} & 116.3\\
This work (\textsc{edge})  & 94.5 & 84.6 & \textbf{89.3} & 91.3 & 78.2 & \textbf{84.3}  & 93.3 & 86.5 & \textbf{89.8} & 89.2 & 80.4 & \textbf{84.6} & 554.0\\
\end{tabular}
}
\vspace{-1mm}
\caption{Complete results on CoNLL-2003.}
\label{tab:conll}
\vspace{-1mm}
\end{table*}

For GENIA we used these features:
\squishenum
\item Words and POS tags (with window of 2 words to the left and right of current word)
\item Words and POS tags n-gram (up to length 4 containing current word)
\item Bag-of-word (with window of 5 words to the left and right of current word)
\item Brown clusters with window of 1 word to the left and right of current word (using 100 or 1000 clusters built from training data only)
\item Word shape (\citet{Finkel:2009})
\item Prefixes and suffixes of current word (up to length 6)
\item Edge type
\squishend


For CoNLL 2003 we used these features:
\squishenum
\item Words and POS tags (with window of 2 words to the left and right of current word)
\item Words and POS tags n-gram (up to length 4 containing current word)
\item Bag-of-word (with window of 5 words to the left and right of current word)
\item Word shape (with window of 2 words to the left and right of current word)
\item Prefixes and suffixes of current word (up to length 5)
\item Orthographic (following \citet{Lu:2015})
\item Edge type
\squishend

The mention penalty feature was added to all models by assigning it to the edges which have the semantics of starting a new entity, similar to how it was defined originally in \citet{Lu:2015}. More specifically, for linear-chain CRF models we add the mention penalty feature to all incoming edges of the {\bf B} and {\bf U} nodes. For our \textsc{State}-based model we add it to all incoming edges of the nodes that includes \texttt{S}. For our \textsc{Edge}-based model we add it to the incoming edges of the $\mathbf{I}$ nodes.

\begin{table*}[ht!]
\small
\centering
{\def\arraystretch{1.1}\tabcolsep=4.5pt
\setlength\doublerulesep{1pt}
\begin{tabular}{lc|c|ccc|ccc|ccc|ccc|ccc}
& & \multirow{2}{*}{\%} & \multicolumn{3}{c|}{\scriptsize{LCRF (single)}} & \multicolumn{3}{c|}{\scriptsize{LCRF (multiple)}} & \multicolumn{3}{c|}{{\scriptsize{\citet{Lu:2015}}}} & \multicolumn{3}{c|}{\scriptsize{This work (\textsc{State})}} & \multicolumn{3}{c}{\scriptsize{This work (\textsc{Edge})}} \\
& & & $P$ & $R$ & $F_1$ & $P$ & $R$ & $F_1$ & $P$ & $R$ & $F_1$ & $P$ & $R$ & $F_1$ & $P$ & $R$ & $F_1$\\
\hline
\multirow{2}{*}{ACE-2004} & O & 42 & 65.6 & 40.9 & 50.4 & 69.6 & 50.4 & 58.5 & 72.5 & 52.4 & 60.8 & 72.2 & 55.1 & 62.5 & 72.1 & 55.3 & 62.6 \\
						 & \O & 58 & 67.4 & 66.6 & 67.0 & 70.5 & 68.2 & 69.4 & 72.5 & 65.0 & 68.6 & 74.2 & 65.4 & 69.5 & 74.1 & 65.5 & 69.5 \\\hline
\multirow{2}{*}{ACE-2005} & O & 32 & 64.6 & 43.4 & 51.9 & 68.5 & 49.3 & 57.4 & 68.1 & 52.6 & 59.4 & 68.6 & 54.7 & 60.8 & 70.4 & 55.0 & 61.8 \\
						 & \O & 68 & 60.4 & 62.7 & 61.6 & 64.1 & 65.3 & 64.7 & 64.1 & 65.1 & 64.6 & 64.9 & 62.7 & 63.8 & 67.2 & 63.4 & 65.2 \\\hline
\multirow{2}{*}{GENIA}    & O & 24 & 78.3 & 52.6 & 62.9 & 78.0 & 59.2 & 67.3 & 76.3 & 60.8 & 67.7 & 77.0 & 60.0 & 67.5 & 76.5 & 60.3 & 67.4 \\
						 & \O & 76 & 76.6 & 70.6 & 73.4 & 74.7 & 70.7 & 72.7 & 73.1 & 70.7 & 71.9 & 75.2 & 70.6 & 72.8 & 74.8 & 71.3 & 73.0\\
\end{tabular}
}
\vspace{-4mm}
\caption{Results on different types of sentences.}
\label{tbl:result-overlapping}
\vspace{-2mm}
\end{table*}

\section{GENIA Preprocessing}
For GENIA, we used GENIAcorpus3.02p that comes with POS tags for each word \cite{Tateisi:2004}. 
Similar to the problem faced by \citet{Finkel:2009} on JNLPBA dataset, we also find tokenization issues in this corpus. 
As described by \citet{Tateisi:2004}, when a hyphenated word such as {\it IL-2-induced} is partially annotated as an entity  (in this case {\it IL-2}), the POS annotation corpus splits it into two tokens, which when done in test set will leak some information about the presence of entity.
Unlike \citet{Finkel:2009} which tried to match the tokenization during testing, we simply further split all tokens at some punctuations (those matching the regular expression \texttt{[-/,.+]}), while keeping the information that they all originally come from the same word.
This has the advantage of simplifying the tokenization procedure, although it makes the task slightly more difficult due to the higher number of tokens.

Also, to handle the discontiguous entities present in GENIA dataset (mainly due to coordinated entities involving ellipsis), following the approach used by the JNLPBA shared task organizer \cite{Kim:2004}, we consider a group of coordinated entities as one structure. For example, in ``{\it \dots the [T- and B-lymphocytes] count in \dots}'', the entities ``{\it T-lymphocytes}'' and ``{\it B-lymphocytes}'' are annotated as one structure ``{\it T- and B-lymphocytes}''.

\section{Results on CoNLL-2003}
Table \ref{tab:conll} shows the full result of the experiments on CoNLL-2003 dataset, which includes the result in CoNLL-2003 development set, and also the results after optimizing the $F_1$ score. We see that since the precision and recall in LCRF model is already balanced, optimizing the $F_1$ score does not improve much, and even slightly decrease the result in the test set. We do see, however, some slight improvements in other models.

\section{Results on Overlapping and Non-overlapping Sentences}
Table \ref{tbl:result-overlapping} shows the complete scores of each model on the two subsets of the test set: the overlapping ones (O) and the non-overlapping ones (\O). We can see that the edge-based model is quite robust to the amount of overlapping mentions in the dataset. For example, in GENIA dataset where the proportion of overlapping mentions is lower compared to the ACE datasets, it still achieves good results both in the overlapping part and the non-overlapping part.

\section{Hyperparameter}

For each model, we tuned the $l_2$-regularization coefficient $\lambda$ from the values \{0.0, 0.001, 0.01, 0.1, 1.0\}. And for GENIA we additionally tuned the number of Brown clusters used from the values \{100, 1000\}.
Table \ref{tbl:hyperparameter} lists the optimal $\lambda$ for each dataset and model. For GENIA, the optimal Brown cluster size was found to be 1000, except for `This work (\textsc{State})', where the best cluster size is found to be 100.

\begin{table}[ht!]
\small
\centering
{\def\arraystretch{1.075}\tabcolsep=2.75pt
\setlength\doublerulesep{1pt}
\begin{tabular}{l|r|r|r|r}
 & \multicolumn{1}{c|}{ACE'04} & \multicolumn{1}{c|}{ACE'05} & \multicolumn{1}{c|}{GENIA} & \multicolumn{1}{c}{CoNLL}\\\hline
LCRF (single) & 0.1 & 0.01 & 0.1 & 0.001 \\
LCRF (multiple) & 0.001 & 0.0 & 1.0 & 0.001 \\
\citet{Lu:2015} & 0.001 & 0.0 & 1.0 & 0.01\\
This work (\textsc{State}) & 0.0 & 0.001 & 1.0 & 0.001\\
This work (\textsc{Edge}) & 0.001 & 0.001 & 1.0 & 0.001
\end{tabular}
}
\vspace{-4mm}
\caption{The value of $l_2$ regularization parameter that gives the best result in development set.}
\label{tbl:hyperparameter}
\vspace{-4mm}
\end{table}

\bibliography{emnlp2017-mention-separators}
\bibliographystyle{emnlp_natbib}
\vphantom{.}